# Real-time Human Detection Model for Edge Devices

Ali Farouk Khalifa[*,a], Hesham N. Elmahdy[a] , Eman Badr[a,b]

[a] *Faculty of Computers and Artificial Intelligence, Cairo University, Giza, Egypt*
[b] *University of Science and technology, Zewail City of Science and Technology, Giza, Egypt*

**Abstract**

Building a small-sized fast surveillance system model to fit on limited resource devices is a challenging, yet an important task. Convolutional Neural Networks (CNNs) have replaced traditional feature extraction and machine learning models in detection and classification tasks. Various complex large CNN models are proposed that achieve significant improvement in the accuracy. Lightweight CNN models have been recently introduced for real-time tasks. This paper suggests a CNN-based lightweight model that can fit on a limited edge device such as Raspberry Pi. Our proposed model provides better performance time, smaller size and comparable accuracy with existing method. The model performance is evaluated on multiple benchmark datasets. It is also compared with existing models in terms of size, average processing time, and F-score. Other enhancements for future research are suggested.

*Keywords:* Deep learning, Convolutional Neural Networks (CNNs), Human Detection, Raspberry Pi

- This research did not receive any specific grant from funding agencies in the public, commercial, or not-for-profit sectors.

## 1. Introduction

Nowadays, personal and assets security is widely demanded. Thus, surveillance systems importance has greatly increased. With surveillance systems, confidential areas can be monitored such as banks, military buildings, and governmental organizations. Utilizing surveillance systems can alleviate or even prevent crimes, and accidents in driving. They also can help in traffic monitoring. Traditional surveillance systems comprise a camera and monitoring person. Recently, surveillance systems are automated through the adoption of internet or using embedded devices. Internet solution adds other issues to the system such as: connection loss, server failure, and hacks. Applying automated surveillance

[*] Corresponding author.
   *Email addresses:* a.farouk@fci-cu.edu.eg (Ali Farouk Khalifa),
ehesham@fci-cu.edu.eg (Hesham N Elmahdy),
emostafa@zewailcity.edu.eg (Eman Badr)

systems on edge devices is challenging.

Surveillance system consists of two phases: motion detection, and object/human detection. Machine/Deep learning techniques are introduced for object/human detection phase. Most of proposed deep learning approaches are based on Convolutional Neural Networks (CNNs) due to high accuracy achieved. CNN models are a powerful tool for classification and detection tasks. Typically, CNN models are large complex models in order to learn different presentations of the data and hence better accuracy. However, they acquire huge computational resources. Multiple calls for such big models, even on a server, can cause server failure. On the other hand, fitting a CNN model on small devices is a challenging task.

Recently, lightweight models have been adopted [1–4]. Although the proposed models are much smaller than traditional CNN models, they are still relatively too large to fit on small edge devices such as Raspberry Pi. They can be considered fast models compared to traditional ones, but for limited computational resources devices, they might take consider- ably long time and will acquire all the device resources. As a result, the device may fail or not be real-time responsive.

In this paper, we propose a lightweight CNN-based model. Our model can fit on edge devices with limited resources such as Raspberry Pi. Our Proposed model size is small and its response time is fast. We evaluated the proposed model by assessing its performance on different benchmark datasets. We summarize our research contributions as follows:

- The architecture of the proposed Lightweight model is described. Our model comprises conventional convolution along with depthwise/pointwise separable convolutions. Bottleneck layer is added to reduce the feature maps size and consequently the complexity of computations. Dense connected layers are chosen not to be exploited as most of the computations and parameters are in the dense layers.
- The model is trained and tested on benchmark datasets. Utilized datasets are: INRIA, ImageNet, Stanford, PascalVoc. Testing against benchmark datasets is a necessary step to be able to reliably assess the model performance in accordance with other existing models.
- Performance is evaluated in terms of size, processing time, and F-Score as accuracy measures where our model is compared against existing edge device models.

The rest of the paper is organized as follows: section 2 includes a literature review. Section 3 describes the utilized datasets in training and testing. In Section 4, proposed model and its specifications are introduced. Section 5 illustrates the performance of the model on different datasets. We conclude our work with suggestions for future improvements in section 6.

## 2. Literature Review

Monitoring humans is an important task of surveillance systems. Surveillance systems consist of two main steps: motion detection, and human detection. Motion detection is to detect whether there is any moving object in the scene through tracking changes. Many methods were introduced to serve this subject such as: frame difference, optical flow [5], and background subtraction [6]. Performance time and object information provided are the main key factors that distinguish these techniques. After motion is detected and candidate regions that may contain human or objects of interest are extracted, the next step of the system is human detection. Human detection consists of 2 sub-phases: object feature extraction, and detector model / classifier.

Feature extraction is the process of finding descriptors that can identify and discriminate the object of interest from other objects. There are many features range from general object features to handcrafted, object-specific features such as [7–13]. The most widely used feature for human detection tasks is Histogram of Oriented Gradients (HOG) [14] and its variants [11,15–17]. The extracted features are then fed to a model / classifier that learns how to discriminate the object of interest from other objects. The utilized model can be a machine learning and/or mathematical model such as: Bayesian classifier, AdaBoost [18], cascade classifier [17], support vector machines (SVM) [18], and neural networks (NN). For a complete review, please refer to [19].

In [19], a comparison of different recent automated surveillance systems on Raspberry Pi has been conducted. The surveyed systems are applied in different environments, utilized different datasets in training and testing, and different versions of raspberry pi devices were used. Consequently, a unified benchmark dataset and the same Raspberry pi device are utilized. HOG is mostly used along with SVM for surveillance systems on edge devices. In standard HOG, it uses windows of size 16 * 16 pixels each window is divided to 2*2 cells each cell is of size 8*8. These large windows enables capturing more spatial information. With overlap of 50% of the windows this enables overcoming occlusion and clutter problems. Most used SVM kernel is the linear kernel. It has less heavy computations than other kernels such as Polynomial or RBF. Although the F-Scores of these models were promising. The size of the models is large and the execution time is not suitable for real time tasks on edge devices. Notably, these models were trained on images of size 128*64 cropped on human objects. Polynomial or RBF kernels utilization can enhance the results yet will increase the execution time. Exploiting HOG features increases the computations due to resulted feature vector size. Thus, dimensionality reduction methods can be used to produce shorter length feature vector sustaining the information.

Deep learning is a subfield in machine learning that has revolutionary results across many domains recently. Deep learning models usually consist of multiple processing layers to learn different levels of details about the data and its inherited features. With the increasing advancements in computing technologies and processing power, these models are trained with a huge amount of data and are achieving higher accuracy than the state-of-the-art machine learning models. Convolutional Neural Network (CNN) is a neural network that consists of convolutional part before the NN hidden layers. In CNN, there are no features needed to learn the object characteristics as the object features are learned in the convolutional part. Utilizing CNN is recently increased as the features are automatically learned and there would be no need to the handcrafted features that may be not suitable for the object of interest [19].

Various architecture of CNNs have been developed to solve different complex problems in computer vision such as object detection. Some examples are: AlexNet [20], VGG [21], Single Shot MultiBox Detector (SSD) [22], Res-Net [23], Inception [24], and many others. Typically, these are very large models that need long time to be trained. For example, AlexNet took 5 to 6 days to train on two NVIDIA GTX 580 3GB GPUs. VGG also took 2 to 3 weeks to train on a system that has 4 NVIDIA Titan Black GPUs. Moreover, the resulted model needs large memory to be used in testing. Consequently, applying those models requires extensive computational resources and, typically, high costs.

These different architectures were introduced over the years with the rise of powerful GPUs. They differ in accuracy and speed. Many modifications of the CNN architecture were introduced such as: skip connections or inception block presented in ResNet and Inception models respectively. These modifications enable the models to learn the objects features better and tremendously enhance the discrimination between objects. Additionally, these modifications improve the accuracy; nevertheless, this came on the expense of model size and speed.

However, there is a critical need to use such models on small devices such as: mobiles, edge devices in IoT systems, or cameras in surveillance systems. Deep learning models also cannot run on small devices, as a result, servers and internet connections are needed which introduces additional problems. For example, if the server is down or the internet connection is lost, the sys-tem would fail. This can be crucial if the system is used for critical purposes such as monitoring system in: banks, governmental buildings, or military areas [19]. Thus, there is a need to develop lightweight CNN (LW-CNN) models that are efficient enough to acquire low computations and low memory. Architectures such as: MobileNet [1], SqueezeNet [2], ShuffleNet [3], and L-CNN [4], are designed such that they are optimized in computation and small in size compared to state-of-the-art models. However, for edge devices the model response will not be in real-time.

MobileNet [1] is a CNN model that focuses on producing small, optimized in latency networks. Its main purpose was to fasten the execution process. However, this came on the expense of accuracy. It is built on the depthwise separable convolution [25] to reduce computations in convolution layers.

In [4], L-CNN is typically built on the architecture of MobileNet. The main difference between L-CNN and MobileNet is in the number of filters utilized in each layer. L-CNN can utilize either a softmax classifier after the convolu-tional layers or regression layers that are used in determining the bonding boxes around the detected objects. Table 1, shows a comparison and the differences between MobileNet and L-CNN architectures.

In [2], SqueezeNet is introduced. SqueezeNet is typically consists of a Fire Module that consists of two convolutional layers: squeeze layer and expand layer, Fig. 1. Squeeze layer consists of multiple 1*1 filters only. Expand layer consists of a mix of 1*1 and 3*3 filters.

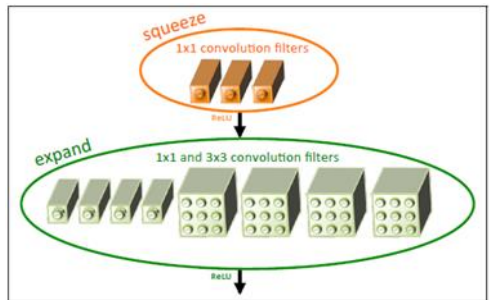

Fig. 1: Fire Module [2].

| Type / Window stride | MobileNet | | L-CNN | |
|---|---|---|---|---|
| | Filter Shape | Input Size | Filter Shape | Input Size |
| **Conv / s2** | 3 * 3 * 3 * 32 | 224 * 224 * 3 | 3 * 3 * 3 * 16 | 224 * 224 * 3 |
| **Conv dw / s1** | 3 * 3 * 32 dw | 112 * 112 * 32 | 3 * 3 * 16 dw | 112 * 112 * 16 |
| **Conv / s1** | 1 * 1 * 32 * 64 | 112 * 112 * 32 | 1 * 1 * 16 * 16 | 112 * 112 * 16 |
| **Conv dw / s2** | 3 * 3 * 64 dw | 112 * 112 * 64 | 3 * 3 * 16 dw | 112 * 112 * 16 |
| **Conv / s1** | 1 * 1 * 64 * 128 | 56 * 56 * 64 | 1 * 1 * 16 * 32 | 56 * 56 * 16 |
| **Conv dw / s1** | 3 * 3 * 128 dw | 56 * 56 * 128 | 3 * 3 * 32 dw | 56 * 56 * 32 |
| **Conv / s1** | 1 * 1 * 128 * 128 | 56 * 56 * 128 | 1 * 1 * 32 * 32 | 56 * 56 * 32 |
| **Conv dw / s2** | 3 * 3 * 128 dw | 56 * 56 * 128 | 3 * 3 * 32 dw | 56 * 56 * 32 |
| **Conv / s1** | 1 * 1 * 128 * 256 | 28 * 28 * 128 | 1 * 1 * 32 * 64 | 28 * 28 * 32 |
| **Conv dw / s1** | 3 * 3 * 256 dw | 28 * 28 * 256 | 3 * 3 * 64 dw | 28 * 28 * 64 |
| **Conv / s1** | 1 * 1 * 256 * 256 | 28 * 28 * 256 | 1 * 1 * 64 * 64 | 28 * 28 * 64 |
| **Conv dw / s2** | 3 * 3 * 256 dw | 28 * 28 * 256 | 3 * 3 * 64 dw | 28 * 28 * 64 |
| **Conv / s1** | 1 * 1 * 256 * 512 | 14 * 14 * 256 | 1 * 1 * 64 * 128 | 14 * 14 * 64 |
| **5* conv dw / s1** | 3 * 3 * 512 dw | 14 * 14 * 512 | 3 * 3 * 128 dw | 14 * 14 * 128 |
| **5* conv / s1** | 1 * 1 * 512 * 512 | 14 * 14 * 512 | 1 * 1 * 128 * 128 | 14 * 14 * 128 |
| **Conv dw / s2** | 3 * 3 * 512 dw | 14 * 14 * 512 | 3 * 3 * 128 dw | 14 * 14 * 128 |
| **Conv / s1** | 1 * 1 * 512 * 1024 | 7 * 7 * 512 | 1 * 1 * 128 * 256 | 7 * 7 * 128 |
| **Conv dw / s2** | 3 * 3 * 1024 dw | 7 * 7 * 1024 | 3 * 3 * 256 dw | 7 * 7 * 256 |
| **Conv / s1** | 1 * 1 * 1024 * 1024 | 7 * 7 * 1024 | 1 * 1 * 256 * 256 | 7 * 7 * 256 |
| **Avg Pool / s1** | Pool 7 * 7 | 7 * 7 * 1024 | Pool 7 * 7 | 7 * 7 * 256 |
| **FC / s1** | 1024 * 1000 | 1 * 1 * 1024 | Regression layers | 1 * 256 |
| **Softmax / s1** | Classifier | 1 * 1 * 1000 | Classifier | 1 * 256 |

*Table 1: Comparison between MobileNet and L-CNN architectures.*

SqueezeNet consists of ten layers: conventional convolutional layer followed by eight fire layers and then a final convolutional layer.

As a result, MobileNet, L-CNN, and SqueezeNet reduced the complexity of time by adapting the depthwise separable convolutional layers instead of conventional convolutional layer, and by utilizing small filters sizes and small channels in fire module respectively. Worth to mention, L-CNN is smaller in size than MobileNet where L-CNN footprint in memory is 122 MB as reported in [4].

## 3. Dataset

Three datasets are utilized in our work: INRIA [14], Stanford human actions dataset [26] and ImageNet dataset [27]. INRIA data set consists of two classes: human, and non-human. There are 614 images that contain humans and 1218 person-free images are utilized for training. For testing, 288 human images and 453 non-human images are utilized.

Stanford dataset contains images for human actions such as: clapping, blowing bubbles, brushing teeth, jumping, climbing, etc. Image name indicates its label. Images are selected manually such that the focus on human not on action. For example: a clapping action class images are mostly focused on the hands not the human. We use 656 images from Stanford dataset where all are for human class.

ImageNet dataset is a large dataset that consists of millions of high-resolution images collected from the web. ImageNet Large Scale Visual Recognition Challenge (ILSVRC) is an image classification annual challenge. In ILSVRC, there are 1.2 million high-resolution images categorized to 1000 category. We utilize 833 images from Person classes positive class. Images from other classes were manually selected randomly as non-human class (846).

PascalVoc [28] is a widely used dataset. It consists of a total of 9963 images. There are 5011 images for training and 4952 for testing. It includes images for 20 different objects such as: person, bird, car, chair, etc. We utilize 207 images that contain human for testing.

We used a total of 4167 human and non-human samples from the three datasets: INRIA, Stanford, and ImageNet. We used 3966 samples in training phase, 2003 images are for human and 1963 human-free, and 201 for validation. For testing, we utilize 794 human images from INRIA, ImageNet, and PascalVoc [28] testing sets. For non-human, 752 images from INRIA and ImageNet testing set are utilized.

## 4. Methodology

### 4.1. Overview of the proposed model

A lightweight deep CNN model that efficiently utilizes re-sources on embedded devices is proposed. The model consists of 10 layers. Both conventional and depthwise separable convolutions are utilized. The first three layers are conventional convolutional layer the rest of the layers are depthwise separable convolutional layers. Utilizing depthwise separable convolution improves processing time and makes the model smaller. The 9th layer is a bottleneck convolutional layer. The output of the convolutional layers is then flattened and fed to a sigmoid classifier. Dense layers are not utilized for the sake of reducing model size and processing time as most of the computations and parameters appear in dense layers. Fig. 2, shows the architecture of our proposed model.

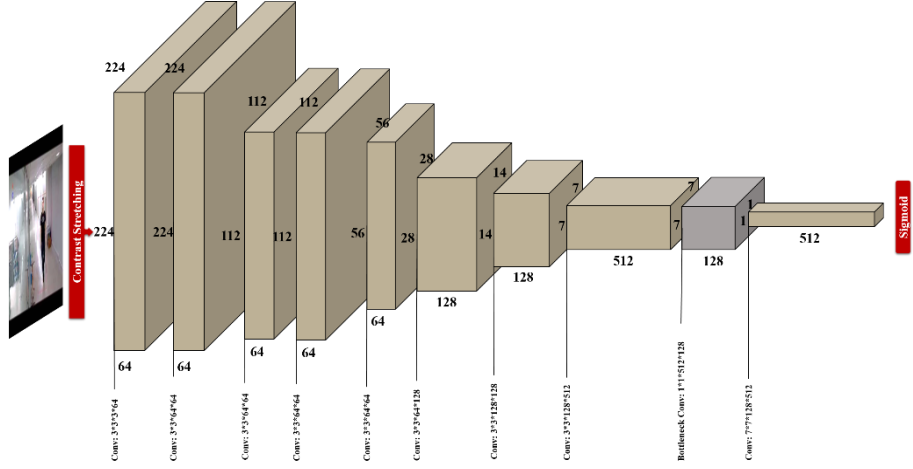

Fig. 2: Proposed model architecture.

*4.2. Conventional and Separable convolutions*

LW-CNN models' main goal is to reduce the computations needed in the convolution part of the CNN. Typically, the conventional 2D convolution is converted to depthwise separable convolution that gives the same results of 2D convolution but with dramatic reduction in the parameters size and computation complexity. Each convolutional layer is converted to a depthwise separable layer which is consisted of 2 layers: depthwise and pointwise convolutional layers. Depthwise separable convolutional (DSC) layer is first introduced in [25].

Fig. 3, demonstrates the conventional convolutional layer. The input tensor F of the conventional convolutional layer has dimensions of $D_f*D_f*M$ that are (height, width, channels). The conventional convolutional layer has N filters of size $D_k*D_k*M$. As a result, the output of this convolution will be a tensor G of size $D_g*D_g*N$. The $D_g$ of the output tensor is equal to the $D_f$ of the input tensor considering that convolution is applied with padding. Equation 1 shows the complexity of this process.

$$O(n)_{Conv} = D_k * D_k * M * N * D_f * D_f \quad (1)$$

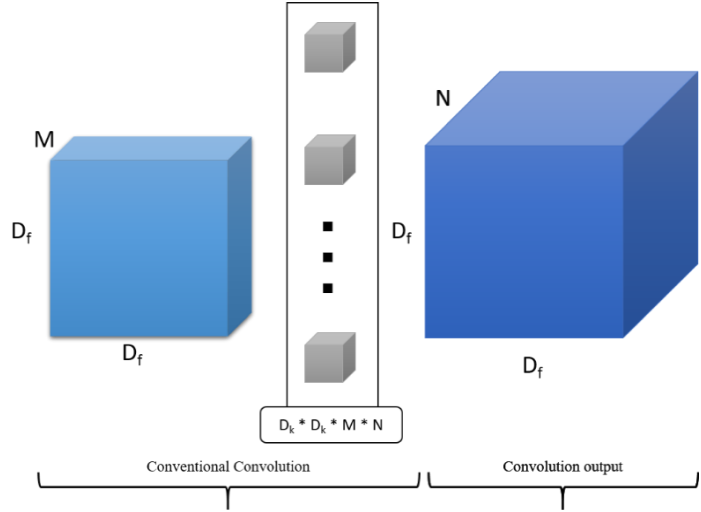

Fig. 3: Conventional Convolutional Layer.

On the other hand, DSC contains two layers: depthwise and pointwise layers. As shown in Fig. 4, the first depthwise layer is to convolve the input tensor F of M channels with M filters of size $D_k*D_k*1$, one filter per channel. The result of this convolution is M tensors each of size $D_f*D_f*1$. The second pointwise convolution layer convolves the output of depthwise layer with N filters each of size $1*1*M$. As a result, the final output tensor is of size $D_f*D_f*N$ which has the same size as the conventional convolutional layer output tensor but with less computations as shown in Equation 2 that represent the computation complexity of DSC layer.

$$O(n)_{ConvDepthwisePointwise} = (D_k * D_k * M * D_f * D_f) + (N * M * D_f * D_f) \qquad (2)$$

Equation 3, compares the complexity of both layers conventional convolutional and depthwise separable and the reduction in time can be noticed through the division of the complexities.

$$\text{Ratio} = \frac{(D_k * D_k * M * D_f * D_f) + (N * M * D_f * D_f)}{D_k * D_k * M * N * D_f * D_f} = \frac{1}{N} + \frac{1}{D_k^2} \qquad (3)$$

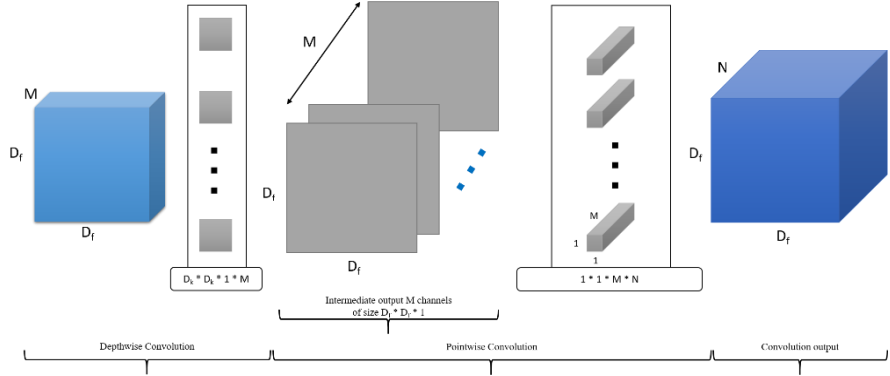

Fig. 4: Depthwise separable convolutional layer.

*4.3. Preprocessing*

We utilize multiple datasets that are collected in different situations (refer to dataset section). Consequently, this may cause different distribution of intensity levels over the images. To overcome this problem, contrast stretching is utilized as a preprocessing step before training and testing. Contrast stretching enhances the image contrast through stretching the image pixels values to span the full intensity range.

*4.4. Kernel Sizes*

Usually deeper models gives better accuracy. Kernels size can affect the learning process as larger kernels in earlier layers can capture more spatial information. However, developing large kernels with deep models will lead to large model size and slow execution.

In the proposed architecture, all layers have filters of size 3*3 except the bottleneck layer and the last convolutional layer. Each layer in the first five layers has 64 filters. Each of the 6[th] and 7[th] layers has 128 filters and the 8[th] and 10[th] layers have 512 filters each. Layer 9 is the bottleneck layer and it has 128 filters of size 1*1. The last layer filters size is 7*7 which is designed to be the same size as the input tensor to this layer. Accordingly, a 1D vector is resulted to reduce the parameters produced from the final layer of the model. Table 2 shows the proposed model specifications.

The larger the receptive field, the more detected spatial features. Hence, model performance is increased through rich information extraction especially in earlier layers. The receptive field of the whole network or part of it can be computed using equation 4. Where for a layer k, $R_k$ is the effective receptive field of this layer, $f_k$ is the kernel size, and $s_i$ is the stride applied in the filters [29]. Consequently, the receptive field of the first three layers in the proposed model is equivalent to one layer of 7*7 kernel size. Additionally, the 4th and 5th layer receptive field is equivalent to one layer of 5*5 kernel size. As a result, the model integrates large kernel sizes implicitly along with deeper layers. Fig. 5 visually presents how 5-point convolution is equivalent to two sequential 3-point convolution. Hence, large odd size kernel convolution can be converted to multiple sequential 3*3 convolutions [30].

$$R_k = R_{k-1} + (f_k - 1) \prod_{i=1}^{k-1} s_i \quad (4)$$

| Type / Window stride | Proposed Model | |
| --- | --- | --- |
| | **Filter Shape** | **Input Size** |
| **Conv / s1** | 3 * 3 * 3 * 64 | 224 * 224 * 3 |
| **Conv / s1** | 3 * 3 * 3 * 64 | 224 * 224 * 3 |
| **Conv / s1** | 3 * 3 * 3 * 64 | 224 * 224 * 3 |
| **Conv dw / s1** | 3 * 3 * 64 dw | 112 * 112 * 64 |
| **Conv / s1** | 1 * 1 * 64 * 64 | 112 * 112 * 64 |
| **Conv dw / s1** | 3 * 3 * 64 dw | 112 * 112 * 64 |
| **Conv / s1** | 1 * 1 * 64 * 64 | 112 * 112 * 64 |
| **Conv dw / s1** | 3 * 3 * 128 dw | 56 * 56 * 128 |
| **Conv / s1** | 1 * 1 * 128 * 128 | 56 * 56* 128 |
| **Conv dw / s1** | 3 * 3 * 128 dw | 28 * 28 * 128 |
| **Conv / s1** | 1 * 1 * 128 * 512 | 28 * 28 * 128 |
| **Conv dw / s1** | 3 * 3 * 512 dw | 14 * 14 * 128 |
| **Conv / s1** | 1 * 1 * 128 * 512 | 14 * 14 * 128 |
| **Bottleneck / s1** | 1 * 1 * 512 * 128 | 7 * 7 * 512 |
| **Conv dw / s1** | 7 * 7 * 128 | 7 * 7 * 128 |
| **Conv / s1** | 1 * 1 * 128 * 512 | 1 * 1 * 128 |
| | **Flatten** | |
| **Softmax / s1** | Classifier | 3 * 3 * 512 |

*Table 2. Proposed model's layer specifications.*

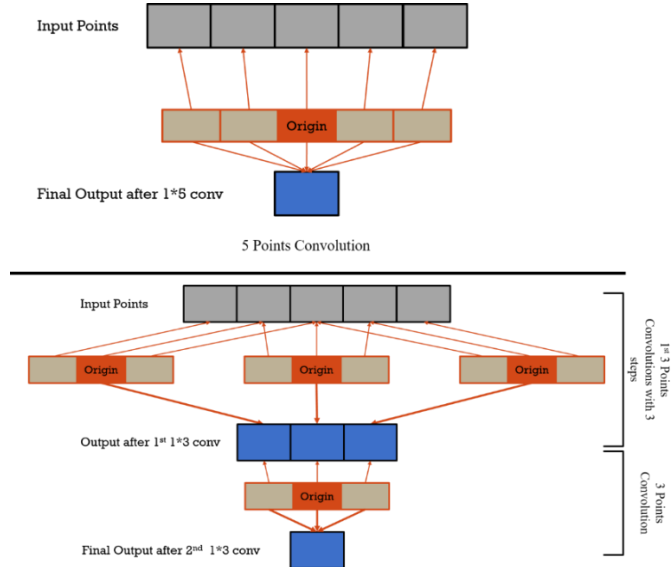

Fig. 5: (1*5) Convolution and its equivalent double (1*3) convolution.

## 4.5. Bottleneck Layer

Bottleneck layer [29] is a convolutional layer in which the kernels dimensions are 1*1. The number of 1*1 filters used in this layer controls the dimension of the output feature maps. Typically, if the number of filters is smaller than the previous and the successive layers. It will result in reduction in the feature maps number while sustain the intrinsic features. Utilizing bottleneck layers increases the number of parameters a little but much improves the processing time. Fig. 6 illustrates the bottleneck convolution effect.

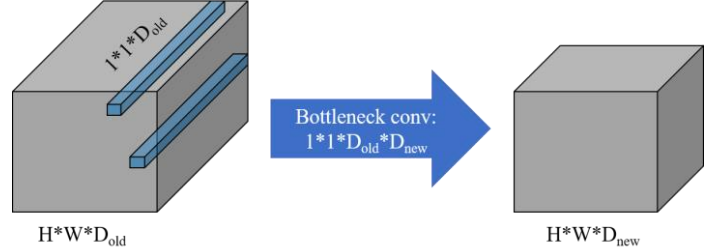

Fig. 6: Bottleneck illustration graph. H and W are the feature map height and width respectively. $D_{old}$ and $D_{new}$ are the number of input feature map channels and the resulted feature map channels respectively

### 4.6. Weight Initialization

Weight initialization is a vital step in training the model. Wrong initialization may stall the learning process and it may never converge. Our model is initialized with weights from VGG 16 weights that is trained on ImageNet [27] dataset.

### 4.7. Model Training Details

Size reduction can be achieved through the network by utilizing filter strides in convolutional layers and pooling Layers. Max Pooling is adopted after layers (3, 5, 6, 7, and 8). Pooling kernel size is 2 and with stride of 2 to apply it on non-overlapping regions. Max pooling reduces the computational cost. Rectified Linear Unit (ReLU) activation function is utilized for all layers. ReLU typically converges faster than other activation functions. Its derivatives computations are less heavy than other activations. Both pooling and ReLU add non-linearity to the model to overcome overfitting.

Dropout is added after many layers for generalization. Keeping probability of 0.35 is empirically chosen. Batch Normalization and weight penalization with L2 norm are used also to generalize the model to fit well on unseen data. Batch Normalization adds slight regularization effect that helps in model generalization, but its main effect is to speed up learning process. Max pooling, and dropout are not adopted in bottleneck and last layer. In addition, bottleneck layer does not have batch normalization. Model specifications can be seen in Table 3.

| Filters strides | Stride = 1 |
|---|---|
| Weights initialization | Initialized with VGG 16 trained on ImageNet |
| Padding | Same, Last layer: Valid |
| Epochs | 100 |
| **Batch size** | 40 |
| **Optimizer** | Adam |
| **Learning Rate** | 0.0007 |
| **Penalty multiplier** | L2, 0.001 |

*Table 3. Proposed model specification details.*

## 5. Discussion and Results

Different datasets are utilized to evaluate the generalization of our pro-posed model. The employed images are originated in different situations and environments. Accordingly, contrast stretching is used as a preprocessing step. As a result, too dark or too bright images problem is resolved. Our model is trained on three different datasets: INRIA [14], ImageNet [27] and Stanford human actions [26] datasets to ensure diversity and generality. We tested our model on the following datasets: INRIA test dataset, ImageNet, and PascalVoc [28]. Table 4 illustrates our model performance on these test datasets.

| | TP (%) | TN (%) | FP (%) | FN (%) | F-Score |
|---|---|---|---|---|---|
| **INRIA** | 90.97 | 90.93 | 9.07 | 9.38 | 0.91 |
| **ImageNet** | 83.28 | 71.57 | 28.43 | 16.72 | 0.79 |
| **PascalVoc** | 82.13 | - | - | 17.87 | - |
| **Total** | **85.64** | **83.11** | **16.89** | **14.36** | **0.85** |

*Table 4. Proposed model performance on test sets.*

The model has the ability to detect if there is human in images with average accuracy of ~ 86%. The model can decide that the input image is human-free with accuracy of 90.93% of true negatives on INRIA test dataset. On the other hand, true negatives on ImageNet is (71.57%). That can be explained by the complex nature of its images as they usually contain many non-mutually exclusive objects. Only human images from PascalVoc test dataset are utilized in model testing. Accordingly, TP and FN only appear in Table 4. We believe that utilizing more human and non-human images in training can improve the model accuracy. To give better insight of the model, total measures are calculated. Thus, total TP, TN, FP, FN, and F-Score are shown in Table 4. Accordingly, the model can detect human-free images with accuracy of 83.11%. The model gives an accepted overall accuracy measured in F-Score of 0.85.

The way the model is saved and then retrieved from memory for inference can affect the performance time significantly as the time needed to load and parse the model file is added to the model processing time. So, typically saving the model and retrieving it in an efficient way reduces its size and execution time. As a result, we utilized FlatBuffer[1] serialization library. FlatBuffer is a cross-platform library created at Google for performance-critical applications such as: games. Serialization is typically done a lot by programs to allocate and access the memory during execution. However, this allocation and access to memory can be done inefficiently. As a result, slower performance happens. On the contrary, FlatBuffer encodes the data, here the model, in a binary file as nested objects of (structs, vectors, strings, tables) in an organized manner using offsets. As a result, the generated binary file can be accessed and traversed in-place (like pointers). Typically, this will lead to fast execution, due to fast decoding process, in addition to smaller file size and hence, faster processing time. Table 5 shows model size and processing time before and after utilizing FlatBuffer.

| Model | Size (MB) | PT (seconds) |
|---|---|---|
| **Without utilizing FlatBuffer library** | 4 | 2.3 |
| **Utilizing FlatBuffer library** | 1 | 0.001 |

*Table 5. Model size and PT before and after utilizing FlatBuffer.*

An ablation study is conducted to show the effect of removing and replacing some layers on the model performance in terms of accuracy, size, and execution time. In this study, the first three convolution layers that each has kernel size of 3*3 of the model are replaced with one layer with kernels of size 7*7. Also, the 4$^{th}$ and 5$^{th}$ layers are replaced with one layer with kernel size of 5*5 and the rest of the model is kept the same. The number of layers in the model now is reduced from 10 to 7 and the filter sizes in the first two layers now are larger, but the receptive field of the network is the same. The average processing time is reduced to 1 second, without applying FlatBuffer. Consequent to reducing the number of layers to 7 layers, the number of parameters is reduced. As a result, the total F-Score of the model is reduced to 0.83 which is 0.02 lower than the original model. The model size is reduced as well to 3 MB. So, as a summary, this modification resulted in faster and smaller model and slight reduction in the accuracy.

| Type / Window stride | Filter Shape | Input Size |
|---|---|---|
| **Conv / s1** | 7 * 7 * 3 * 64 | 224 * 224 * 3 |
| **Conv dw / s1** | 5 * 5 * 64 dw | 112 * 112 * 64 |
| **Conv / s1** | 1 * 1 * 64 * 64 | 112 * 112 * 64 |
| **Conv dw / s1** | 3 * 3 * 64 dw | 56 * 56 * 64 |
| **Conv / s1** | 1 * 1 * 64 * 128 | 56 * 56 * 64 |
| **Conv dw / s1** | 3 * 3 * 128 dw | 28 * 28 * 128 |
| **Conv / s1** | 1 * 1 * 128 * 128 | 28 * 28* 128 |
| **Conv dw / s1** | 3 * 3 * 128 dw | 14 * 14* 128 |
| **Conv / s1** | 1 * 1 * 128 * 512 | 14 * 14* 128 |
| **Bottleneck / s1** | 1 * 1 * 512 * 128 | 7 * 7 * 512 |
| **Conv dw / s1** | 7 * 7 * 128 | 7 * 7 * 128 |
| **Conv / s1** | 1 * 1 * 128 * 512 | 1 * 1 * 128 |
| **Flatten** | | |

---

[1] https://google.github.io/flatbuffers/

| Softmax / s1 | Classifier | 1 * 1 * 512 |

*Table 6 The model after replacing the layers.*

In [19], INRIA dataset is utilized as a benchmark to evaluate the accuracy and performance of different models reported in the literature. We utilize the same dataset to evaluate our model against previous models. Accordingly, we modified the proposed model input to accept images of size 128*64 instead of 224*224. The model is trained and tested on INRIA dataset.

Table 7 contains the evaluation and comparison results. It investigates the accuracy in terms of F-score, the average processing time (PT), and the models size. It shows the dramatic reduction in model size as well as the speed of the proposed model over the SVM models. The proposed model outperforms the SVM models in terms of F-Score which is 0.95 while best SVM model F-Score is 0.93.

|  | Size (MB) | PT (s) | F-Score |
|---|---|---|---|
| **RBF SVM + SW2S** | 120 | 2.11 | 0.84 |
| **RBF SVM + standard HOG with SVM** | 120 | 1.77 | 0.85 |
| **Polynomial SVM + SW2S** | 35 | 1.57 | 0.84 |
| **Polynomial SVM standard HOG with SVM** | 35 | 1.51 | 0.92 |
| **Linear SVM + SW2S** | 30 | 1.55 | 0.93 |
| **Linear SVM + standard HOG with SVM** | 30 | 1.49 | 0.93 |
| **Proposed Model** | 1 | $1*10^{-3}$ | 0.95 |

*Table 7. Models trained on INRIA dataset comparison.*

Table 8 shows detailed performance results of the model against the SVM models. The model ability to detect human-free images is better than SVM models with TN of 95.16%. Moreover, the model detects humans with a decent TP percentage of 94.74%. Accordingly, the model can do the balance between TP and TN detection better than the SVM models. As a result, the proposed model is performing better than SVM models in terms of F-Score accuracy measure, size, and execution time.

|  | TP (%) | TN (%) | FP (%) | FN (%) |
|---|---|---|---|---|
| Linear SVM + standard HOG with SVM | 93.209 | 92.5 | 7.5 | 6.791 |
| Linear SVM + SW2S | 94.907 | 91.103 | 8.8965 | 5.093 |
| **Proposed Model** | **94.74** | **95.16** | **4.84** | **5.26** |
| Polynomial SVM standard HOG with SVM | 94.567 | 88.135 | 11.865 | 5.433 |
| Polynomial SVM + SW2S | 97.963 | 64.310 | 35.6898 | 2.037 |
| RBF SVM + standard HOG with SVM | 95.925 | 71.052 | 28.948 | 4.075 |
| RBF SVM + SW2S | 98.132 | 65.475 | 34.525 | 1.868 |

*Table 8. Proposed Model performance results on INRIA dataset against SVM models in* [19].

To evaluate our proposed model with other lightweight models, a comparison to MobileNet and LCNN has been established. Summarizing the results in Table 9, it is noticeable that MobileNet and LCNN higher accuracy than our model in detecting human existence (TP) by 5-6%. On the other hand, the proposed model can detect human free images (TN) better than both MobileNet and LCNN by 8-9% on average. Consequently, the proposed model has the highest total F-Score. Further, our model size is much lower than MobileNet. Although LCNN size is smaller than the proposed model, the memory footprint of the proposed model is lower than LCNN. Utilizing the memory profiler application utilized in LCNN [4] paper. L-CNN has memory footprint of 122 MB as reported while our model has memory footprint of 27.54 MB which indicated that our model utilizes memory more efficiently.

| | TP (%) | TN (%) | FP (%) | FN (%) | F-Score | Size (MB) |
|---|---|---|---|---|---|---|
| **Proposed Model** | 0.85 | **0.83** | 0.17 | 0.14 | **0.85** | 4 |
| **MobileNet** | 0.90 | 0.75 | 0.25 | 0.10 | 0.84 | 37 |
| **L-CNN** | **0.91** | 0.74 | 0.26 | 0.09 | 0.84 | 3 |

*Table 9 Proposed model vs. MobileNet vs. LCNN results on test sets.*

## Environment Details

Google Colab[2] environment is utilized to train our model. Google colab is a Linux environment that provides variety of processing sources such as: CPU, GPU, or TPU. We utilized the provided Tesla K80 GPU. This GPU has 2496 CUDA cores and 12GB GDDR5 VRAM. We used Python programming language and Keras machine learning API using Tensorflow as a backend. Raspberry Pi 3 model B with the following specs: 1 GB RAM, quad-core 64-bit ARM Cortex-A53 processor clocked at 1.2 GHz, Broadcom VideoCore IV of 400MHz GPU was utilized.

## 6. Conclusions and future directions

Different architectures of CNNs have been introduced in recent years. Despite the significant accuracy of these models, size and processing time are vital limitations to fit on limited resource devices. SVM based models are used to be the most widely used on edge devices, yet they are still large and have a relatively long processing time. A reliable small-sized and fast CNN based model that achieves comparable accuracy and can meet limited resources devices requirements is proposed. Adapting techniques such as a bottleneck layer and receptive fields enhances the reported accuracy, average processing time, and size of the model.

---

[2] https://colab.research.google.com/

The proposed model achieves 0.85 F-Score on multiple datasets. The model has small size and very low response time. On INRIA dataset, the best SVM model gives F-Score of 0.93 while our proposed architecture F-Score is 0.95 with huge reduction in model size and processing time. Additionally, the proposed model is compared against some state-of-the-art lightweight models such as MobileNet and LCNN. The proposed model outperforms MobileNet and LCNN in terms of total F-Score. Our model is much smaller in size than MobileNet and has a small memory footprint than LCNN. Utilizing more images from different dataset in training can enhance the model generality and effectiveness.

There is a room for more enhancements. Adapting other modifications in the CNN architecture such as: skip connections or Inception blocks can improve the model accuracy yet may increase the size.

Because of its small size, fast real time response, and efficient utilization of memory, our model can open the door for different applications that can benefit from such architecture where computational resources are limited such as any edge device as well as mobile devices. Utilizing appropriate data, our model can be tuned and used in many applications such as: autonomous driving, human activity detection, and health related applications, smart homes, disabled assistance, etc. It can be also used in person-specific applications that require small amount of images for specific purpose.